\pdfoutput=1

\documentclass[11pt]{article}

\usepackage{acl}

\usepackage{times}
\usepackage{latexsym}

\usepackage[T1]{fontenc}

\usepackage[utf8]{inputenc}
\usepackage{dblfloatfix}
\usepackage{microtype}
\usepackage{graphicx} 
\usepackage{url}
\title{SynSciPass: detecting appropriate uses of scientific text generation}

\author{Domenic Rosati \\
  scite / Brooklyn, NY \\
  \texttt{dom@scite.ai}}

\begin{document}
\maketitle
\begin{abstract}
Approaches to machine generated text detection tend to focus on binary classification of human versus machine written text. In the scientific domain where publishers might use these models to examine manuscripts under submission, misclassification has the potential to cause harm to authors. Additionally, authors may appropriately use text generation models such as with the use of assistive technologies like translation tools. In this setting, a binary classification scheme might be used to flag appropriate uses of assistive text generation technology as simply machine generated which is a cause of concern. In our work, we simulate this scenario by presenting a state-of-the-art detector trained on the DAGPap22 with machine translated passages from Scielo and find that the model performs at random. Given this finding, we develop a framework for dataset development that provides a nuanced approach to detecting machine generated text by having labels for the type of technology used such as for translation or paraphrase resulting in the construction of SynSciPass. By training the same model that performed well on DAGPap22 on SynSciPass, we show that not only is the model more robust to domain shifts but also is able to uncover the type of technology used for machine generated text. Despite this, we conclude that current datasets are neither comprehensive nor realistic enough to understand how these models would perform in the wild where manuscript submissions can come from many unknown or novel distributions, how they would perform on scientific full-texts rather than small passages, and what might happen when there is a mix of appropriate and inappropriate uses of natural language generation.
\end{abstract}

\section{Introduction}

\begin{table}[]
\begin{tabular}{llll}
\hline
 & DAGPap22 & SynSciPass & Scielo \\ \hline
DAGPap22 & 99.6 & 31.4 & 52.0 \\
SynSciPass & 81.3 & 98.6 & 65.6 \\ \hline
SciBERT & 98.3 &  &  \\
TF-IDF & 82.0 &  &  \\ \hline
\end{tabular}
\caption{F1 scores on the DAGPap22, SynSciPass, and Scielo datasets including baselines for DAGPap22 (see Appendix \ref{models} for model details)}
\label{tab:kaggle}
\end{table}

While estimated submission rates of machine generated scientific papers are still small \citep{cabanac_prevalence_2021}, contemporary text generation models can generate highly fluent scientific text \citep{generative_pretrained_transformer_can_2022} and manuscripts constructed this way could easily be produced en masse potentially introducing an unprecedented threat to scientific publishing and research integrity. Despite this risk, machine generated text in scientific settings have appropriate uses such as with assistive technology like translation, paraphrasing, and speech-to-text \citep{li_generating_2022}. \footnote{This is not to say that other malicious applications of these technologies such as disguising plagarism do not exist or that use of poor quality text generation technologies don't introduce problems such as nonsensical phrases (see \citet{cabanac_tortured_2021})} Scientific manuscripts may increasingly use both appropriate and inappropriate text generation technologies. If appropriate uses of text generation cause a manuscript to be flagged or rejected this could harm populations that might already struggle with manuscript writing and submission. For instance, even if publisher's intention is only to guide editors, misclassified manuscripts can unintentionally bias editors decisions. Inspired by \citet{schuster_limitations_2020}, we ask whether we can develop a method that could adequately distinguish between appropriate and inappropriate uses of text generation by identifying the category of tool being used such as for translation or paraphrase.

Alarmingly, our study finds that a DeBERTa v3 \citep{he_debertav3_2021} detector that achieves state-of-the-art performance when finetuned on a dataset designed for detecting generated academic text (DAGPap22 \citet{kaggle_detecting_2022}) does poorly on flagging machine generated text under realistic scenarios of appropriate text generation (see Table~\ref{tab:kaggle} which shows SciBERT and logistic regression with TF-IDF baselines trained on DAGPap22 as well as DeBERTa v3 trained on DAGPap22 and SynSciPass). Since misflagging a manuscript as machine generated is harmful to the submitting author, we reframe the problem as detecting the type of tool used for generating text so that authors and publishers can have a more nuanced and neutral approach to understanding flagged texts and guiding editorial decisions. We develop a framework to generate academic texts including labels of the type of technology being used resulting in our dataset of synthetic scientific passages (SynSciPass). Section \ref{dataset} explores how this dataset was constructed and how it could be extended to further improve robustness under domain shifts. In section \ref{reframing}, we show training on SynSciPass results in being able to distinguish the type of technology and how our reframed task helps us move beyond brittle attribution tasks that rely on having access to particular models or the less informative and potentially misleading binary detection task. Finally in section \ref{scielo}, we show that while models trained on our dataset are able to improve robustness under domain shifts for machine generated scientific texts, models for detecting machine generated scientific text are far from ready for safe use by publishers. We provide a roadmap for how to close the gap by focusing on realistic dataset construction that is designed to test detectors ability to robustly generalize across domain shifts.

\section{A framework for robust and granular detection datasets}
\label{dataset}

Previous work on detecting machine generated text has focused on attribution of text to particular models \citep{uchendu_authorship_2020, munir_through_2021}. These approaches have shown the utility of having knowledge of the underlying models for text generation since by having access to those models synthetic corpora can be built for the detection of synthetic text \citep{liyanage_benchmark_2022}. However those approaches are limited to attributions on specific models trained on particular datasets and do not present a realistic or comprehensive scenario where models may be trained on different datasets or models might be unknown. Our framework improves upon model attribution methods by creating corpora from a variety of distributions with a hierarchy of labels including parent labels based on the type of tool used such as for paraphrasing, translation, or novel text generation. By having access to the type of tool used, we are able to make more sophisticated judgments about machine generated text such as allowing translation and paraphrase as appropriate uses of text generation while requiring more scrutiny for fully generated passages. Our framework consists of (1) proposing a taxonomy of approaches, model families, and models with a variety of pretraining or finetuning datasets that might be used for text generation and (2) sampling machine generated text from each model in the taxonomy so that each text can be labeled with a granular labeling scheme according to (1). By doing so, we hope to be able to attribute generated text not only to specific models but also model families and types of technology. With these more generic labels we are able to determine if models generalize detection across model families or across approaches used like if an unseen model for translation were to be introduced.

\subsection{SynSciPass}

In order to address these issues we constructed SynSciPass. For our dataset, we theorized three potential sources of machine generated text (1) free-form text generation using generative models like GPT-2 (2) paraphrase models and (3) translation models. While other approaches like speech-to-text or summarization are also likely used in practice, we restricted to the previously mentioned three. We also did not consider the use of multitask models like GPT-3 that are able to use in-context learning to also do paraphrase and translation \citep{brown_language_2020} which future work should follow up on to understand if different uses of the same model can be properly distinguished. For each approach, we selected a variety of models from different model families in order to try to synthesize a distribution of text generations that might be found in manuscripts (as have been identified by \citet{cabanac_tortured_2021} and \citet{cabanac_prevalence_2021}). These included common services a user might have access to like GPT-2, Spinbot, SCIgen (cf. \citet{cabanac_tortured_2021}) and Google translate as well state of the art approaches for each source such as BLOOM for text generation \citep{bigscience_bloom_2022}. For each technology type, we also included at least one model that was trained on a distribution of scientific text. The final dataset consisted of 110,474 passages of which 99,989 (90.5\%) were not synthetic to introduce more realistic class imbalance given the estimation that only a few papers per million are machine generated \citep{cabanac_prevalence_2021}. Please reference Appendix \ref{synscipass} and \ref{models} for construction details and Table \ref{tab:datasetstats} for full details on dataset construction including the models used to generate data and which model family and technology type they belong to.

\section{Reframing synthetic text detection as multi-class classification for understanding appropriate use}
\label{reframing}

\begin{figure}
    \centering
  \includegraphics[width=1.0\linewidth]{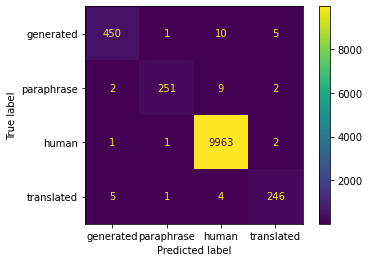}
  \caption{Confusion matrix for multi-class prediction on SciSynPass test set}
  \label{fig:multi_confusion}
\end{figure}

Beyond making models more robust through producing a more comprehensive dataset, our framework reframes binary synthetic text detection as multi-class classification that asks not “Is this passage of text synthetic?” but asks “If this passage is synthetic, how was it created?”. Given that there are several legitimate uses of text generation tools in the scientific writing process such as using assistive technology, in practice this reframing could allow journal editors to make a more nuanced assessment of potentially synthetic text. For example, if a passage of text was detected as using a translation tool, an editor or submitting author can assess if the translation tool was adequate in conveying meaning or if professional translation services should be employed during revisions. If a paraphrase tool was used, editors can assess whether it might have been used to disguise plagiarism or be the result of a poor quality tool such as Spinbot which is known to introduce non-idiomatic phrases \citep{cabanac_tortured_2021}. 

Using this approach we trained a multi-class classifier resulting in a micro-averaged F1 score of 99.6\% (97.4\%, 96.9\%, 99.8\%, and 96.2\% per class F1 for generation, paraphrase, human written, and translation classes respectively) on our held-out test set. To illustrate the models performance we present the confusion matrix in Figure \ref{fig:multi_confusion} showing that our model does quite well across classes even with the large class imbalance. Additionally as seen in Table \ref{tab:kaggle}, the model achieves a F1 score of 81.3\% on DAGPap22 which is quite good considering the different domains and notably is about the same as logistic regression with TF-IDF \emph{despite not being trained on the DAGPap22 dataset}. However, this might simply be that DAGPap22 contains a similar underlying distribution. Unfortunately the distribution of DAGPap22 was not known at the time of writing preventing us from providing a nuanced picture of the differences and overlap between DAGPap22 and SynSciPass.

\begin{table*}[]
\centering
\begin{tabular}{lllllll}
\hline
 & BLOOM & distilgpt2 & gpt2-arxiv & gpt2 & SCIgen & average \\ \hline
SynSciPass & \textbf{96.3} & \textbf{100.0} & \textbf{97.9} & 90.7 & \textbf{100.0} & \textbf{97.0} \\
-gpt2-arxiv & 93.5 & 96.6 & \textbf{97.9} & \textbf{94.4} & \textbf{100.0} & 96.5 \\
-BLOOM & 28.0 & 97.8 & 96.8 & 91.7 & \textbf{100.0} & 82.9 \\ \hline
\end{tabular}
\caption{Ablation study reporting F1 scores on each text generation subset using DeBERTa v3 trained on SynSciPass as a whole, SynSciPass without the samples generated by gpt2-arxiv and SynSciPass without samples generated by BLOOM. See Appendix \ref{synscipass} and \ref{models} for more details.}
\label{tab:ablation}
\end{table*}

In order to see how our multi-class model might generalize across families of text generation models, we performed an ablation study (Table \ref{tab:ablation}) measuring the performance of DeBERTa v3 trained on SynSciPass as a whole, DeBERTa v3 trained on SynSciPass with texts generated by gpt2-arxiv removed and finally DeBERTa v3 trained on SynSciPass with texts generated by BLOOM removed. F1 scores were reported on model performance on each text generate dataset (see Appendix \ref{synscipass} and \ref{models} for details on dataset and model names). In Table \ref{tab:ablation} we see that removing gpt2-arxiv samples results in a small drop in average performance from the model trained on SynSciPass as a whole (96.5 F1 down from 97.0 F1) indicating that \emph{when we test against a seen model trained on a new dataset detectors may still be effective at detecting the type of technology used}. Interestingly removing gpt2-arxiv samples causes the model to do better on gpt2 than SynSciPass as a whole (94.4 F1 up from 90.7 F1). This indicates that having access to the model on a generic domain might be more important than having access to a model pretrained on a specific distribution as has been studied in \citet{rodriguez_cross-domain_2022}. Along these lines we see that removing BLOOM drops performance dramatically on BLOOM from 96.3 to 28.0 F1 score further indicating that having access to underlying models are particularly important and that unseen models may cause detectors to fail. Future work should try to analyze detection models trained to generalize across tools with a wider variety of models including more shifts in underlying pretraining distribution, a variety of model sizes, different sampling procedures, and introducing unseen models.

\section{Out-of-domain Synthetic Passage Detection}
\label{scielo}

\begin{table*}[]
\centering
\begin{tabular}{lllll}
\hline
 & AURC ↓ & F1 ↑ & Precision ↑ & Recall ↑ \\ \hline
DAGPap22 & 47.9 & 52.0 & 49.6 & 54.6 \\
DAPT-TAPT & 49.3 & 57.0 & 50.4 & 65.7 \\
SynSciPass & 51.3 & 65.6 & 50.2 & 94.5 \\
SynSciPass (Translation) & 41.3 & 66.5 & 50.0 & 99.1 \\
SynSciPass (Removed) & 49.6 & 66.5 & 50.1 & 99.1 \\
SynSciPass+DagPap22 & 45.6 & 65.6 & 50.4 & 93.9 \\ \hline
\end{tabular}
\caption{Performance of models presented in Section \ref{scielo} on an out of distribution translation dataset (Scielo Translations) showing a more than 13 point increase over a model trained on DAGPap22 when using SynSciPass.}
\label{tab:oodscielo}
\end{table*}

Following poor performance of models trained only on DAGPap22 on SynSciPass and vice versa (See Table \ref{tab:kaggle}). We wanted to investigate additional domain shifts to understand how robust these models could be in realistic scenarios like seeing new subject domains or new models as this might give us a better picture of how these models might perform in practice. To test robustness over domain shift, we created an additional dataset by sampling human written English passages from Scielo (using \citet{soares_large_2019}) aligned with human written Spanish passages that were translated back into English. This was done to (1) simulate detection where manuscripts might have used translation and (2) simulate where the underlying distribution from Scielo represents a potential stylistic and disciplinary shift from the Pubmed and arXiv domains which have been seen in SynSciPass.

We sampled 1,000 bilingual English-Spanish human written passages from the Scielo bilingual scientific texts dataset \citep{soares_large_2019}. We kept the human written English passages labeled as human generated. Then we translated the aligned human written Spanish passages into English using Google translate and labeled these as machine generated. To get a sense of the resulting lexical overlap between the human and machine translated passages, the BLEU score was 40.9 where the overlap between the English passages with themselves is a BLEU score of 100.0. We tested the resulting dataset of 2,000 passages using (1) DeBERTa v3 trained on DAGPap22 only (DAGPap22) (2) DeBERTa v3 pretrained on the Pubmed split of scientific papers and pretrained on the test and train texts from DAGPap22 and then finetuned on DAGPap22 only (DAPT-TAPT) (3) DeBERTa v3 trained on SynSciPass only (SynSciPass) (4) DeBERTa v3 trained on only translations from SynSciPass (SynSciPass (Translation)) (5) DeBERTa v3 trained with potential confounding factors removed (passages generated by google translate and passages generated by a model finetuned on scielo) SynSciPass (SynSciPass (Removed)) (6) DeBERTa v3 trained on both SynSciPass and DAGPap22 (SynSciPass+DAGPap22) (See Appendix B for full training details). In order to compare the results fairly, we should be clear that SynSciPass uses 1 translation model that was finetuned on the same Scielo dataset \citep{soares_large_2019} to back translate between English and Spanish as well as Google translate to back translate between English and Chinese so there may be some confounding effect of having samples produced by these models. SynSciPass does not contain any samples from Scielo itself. In order to address this potential confounding factor readers should reference the results from SynSciPass (Removed) where both of those sample sets are removed.

In Table \ref{tab:oodscielo}, we see that DAGPap22 does quite poorly with an F1 score of 52\%, mostly due to poor recall indicating that a state-of-the-art model trained on DAGPap22 would perform as if it’s randomly assigning a human generated or machine generated label on translated material mixed in with human written passages from the Scielo domain. Even though this is somewhat expected given that DAGPap22 does not contain information about translations, it is alarming that this is what performance would look like in real life manuscript flagging systems if manuscripts used translators.

A standard approach to improving robustness is pretraining on in-domain and expected task datasets \citep{gururangan_dont_2020}, when utilizing this (DAPT-TAPT) the model does not do too much better (57\% F1) than the one trained on DAGPap22 only. Models trained on SynSciPass do improve (up to 66.5 F1 for SynSciPass (Translation)) but do not perform well enough to be considered safe. These results indicate that common approaches for machine generated text detection are not robust against shifts in domain and result in dismal performance under a realistic scenario. We also measured the area under risk-coverage (AURC) \citep{JMLR:v11:el-yaniv10a} to present what it might be like if we calibrated our models to only select answers they are most confident in. For AURC, DAGPap22 actually does better than SynSciPass and DAPT-TAPT indicating that it’s selective predictions can be made safer. Not surprisingly SynSciPass (Translation) achieves the best AURC of 41.3 indicating that its confidence scores are more meaningful than the others and would perform best at selective prediction, however this requires knowing where the test distribution comes from.

\section{Limitations}

Given the above results it should be clear that machine generated text detectors in the scientific domain are not very robust to realistic domain shifts. While adding nuance to classifications with the multi-class classifier and providing a more comprehensive dataset enables enhanced robustness, the approach is still sensitive to even small shifts in distribution such as using a known model, google translate in the Scielo case, trained on an unseen dataset, Spanish to English scientific passage translation. The major limitation with our framework is that in order to become more robust we will have to continue to collect more distributions to synthesize from and even as we collect a critical mass of potential distributions of machine generated text, our results are inconclusive as to whether models will continue to be more robust to distributions shifts. Our results with BLOOM removed indicate that generalization to unseen text generation models might not be possible with current approaches. Since machine generated text will continue to approach human-level fluency and new approaches will continue to be developed, it will not be tractable to develop a comprehensive dataset that is representative of the underlying distribution of machine generated text. Additionally, since these models are still sensitive to slight shifts in distribution, we suggest that future work should shift focus to improving robustness of detection on out of domain samples such as with selective prediction or more sample efficient approaches of collecting data to become robust as in \citet{rodriguez_cross-domain_2022}. In order to accomplish this, future work should develop a comprehensive suite of tests to evaluate the effects of domain shifts on detectors.

While a multi-class labeling approach might help human evaluators of texts understand why a passage was flagged, this approach should additionally be extended to provide interpretability on why particular passages of texts were flagged. This can be with generating human-like rationales or using methods similar to GLTR \citep{gehrmann_gltr_2019} to assist authors and journal editors in understanding places their manuscript might be improved.

Another limitation of both SynSciPass and DAGPap22 is that they both consist of small passages extracted from scientific texts. Since most manuscripts are submitted as long texts, we are not sure how these results would apply to realistic scientific full-texts, especially when those full-texts include tables, figures, and other non-textual items. While \citet{rodriguez_cross-domain_2022} does provide approaches to address this with passage-based models, future work should still aim to construct datasets that are more realistic and close to the task by providing full-text scientific documents that include layout, figures, and tables. Finally, these datasets should aim to match the extreme class imbalance that has been observed in real world distribution of machine generated texts identified in \citet{cabanac_prevalence_2021}.

\section{Related Works}

\citet{jawahar_automatic_2020} outlines many recent approaches to detecting machine generated text in a variety of domains. The closest to our approach is attribution models that attempt to use a stylometric approach for uncovering the authorship of a text where the author is a particular model or particular model using a particular dataset \citep{jones_are_2022, munir_through_2021}. Our approach is unique in that it focuses on attribution of general classes of tools such as translation, paraphrase, and generation rather than specific models.

While we agree in principal with criticisms of the stylometric approaches that seek to center the veracity and coherence of texts \citep{dou_is_2022, schuster_are_2019}, as text generation models improve, the factuality and fluency gap between machine and human generated text will get smaller and smaller and methods that utilize veracity and coherence will no longer work \footnote{\citet{clark_all_2021} find that humans already cannot reliably distinguish between human and machine generated text produced by GPT-3)}. Additionally, many humans make errors and write poor quality manuscripts so we do not feel like this is a good criterion for detecting machine generated texts but should be clearly separated as an equally important but orthogonal task of understanding the quality of scientific texts. Similarly, we are skeptical of approaches like MAUVE \citep{pillutla_mauve_2021} that rely on distributional artifacts produced by machine generated texts since as text generation models mature the gap between human and machine distributions will also close.

\citet{rodriguez_cross-domain_2022} is the closest to our work in examining the effects of domain shifts in detecting machine generated scientific texts showing that detectors do not generalize well when subject domains shift from physics to biomedicine. While they show that generating even a small number of samples in another domain improves detection, their work is limited to only GPT-2 making their findings reliant on having access to the underlying models. Data augmentation like we used is a common strategy shown to improve the robustness of models in NLP \citep{wang_measure_2022} and is common for examining text generation model attribution in detecting machine generated text since we have access to the underlying text generation models during analysis \citep{uchendu_authorship_2020}. Finally, recent work has examined the robustness of these models \citep{gagiano_robustness_2021, wolff_attacking_2020} but these methods focus on robustness to adversarial attacks such as homoglyphs and misspellings rather than robustness to domain shifts and generalization to unseen models which is studied in this work and which we understand as area with the most promise for both understanding and improving detectors.

\section{Ethics Statement}

The results in this paper should make it clear that at this point machine generated text detectors should not be used in production because they do not perform well on distribution shifts and their performance on realistic full-text scientific manuscripts is currently unknown. Further development is needed on both interpretable and robust detection methods as well as better datasets that are both realistic (such as including full-texts rather than passages) and varied (including comprehensive samples across scientific disciplines). Because erroneously detecting a manuscript as machine generated is a high harm activity, future work should continue more nuanced harm-reduction approaches to synthetic paper detection like the ones introduced in this paper.

\section{Data Availability}

The final constructed dataset, SynSciPass, source code, and models are available at \url{https://github.com/domenicrosati/synscipass}.

\section{Conclusion}

Given our findings, we envision future work along three lines (1) developing machine generated text detectors that are robust across domain shifts and developing realistic datasets that test this robustness comprehensively (2) developing methods of interpretability that help editorial teams detect and manage the use of both appropriate and inappropriate use of text generation models (3) discussion about the safe and ethical application of these technologies and the potential harm involved in their deployment when use cases such as assistive technology are not considered.

We introduced a framework for collecting datasets to improve the robustness and interpretability of detecting machine generated text in the scientific domain. By developing a comprehensive dataset, SynSciPass, we were able to show that models trained on it were not only more robust under domain shifts but also that those models were able to detect the generic type of text generation technology such as for translation, paraphrase, or novel generations which could help understand if a passage was generated by appropriate or inappropriate means. Despite these findings, our work has also shown that current models, including our own, do not perform well in realistic scenarios that change the distribution of text seen. Because of this lack of robustness, we suggest that future work concentrate on formulating both datasets and approaches that comprehensively test machine generated text detectors in a wide variety of realistic and unseen scenarios.

\bibliography{anthology,custom}
\bibliographystyle{acl_natbib}
\appendix

\section{Construction of SynSciPass}
\label{synscipass}

\begin{table*}[]
\centering
\begin{tabular}{llll}
\hline
type & model family & model & passages \\ \hline
generate & bloom & bloom & 1073 \\
 & gpt2 & GPT-2-arxiv\_generate & 998 \\
 &  & distilgpt2 & 998 \\
 &  & gpt2-medium & 998 \\
 & SCIgen & SCIgen & 822 \\
paraphrase & pegasus & pegasus-xsum-finetuned-paws* & 1000 \\
 &  & pegasus-xsum-finetuned-paws-parasci* & 1000 \\
 & spinbot & spinbot & 990 \\
real & real & real & 99064 \\
translate & google\_translate & google\_translate & 901 \\
 & opus & opus-es-en & 794 \\
 &  & opus-es-en-scielo* & 901 \\ \hline
\end{tabular}
\caption{Approaches used for data augmentation and number of passages generated. Models with an asterisk were trained by the authors. Spinbot, SCIgen, and google translate are the names of the services used available online. The rest of the models are or will be made available on the huggingface repository under those names.}
\label{tab:datasetstats}
\end{table*}

SynSciPass was constructed using 100,000 passages that were randomly sampled from the scientific papers dataset \citep{cohan_discourse-aware_2018}. Each passage was between 2 and 10 sentences randomly sampled from the full-text of single publication from both the arXiv and Pubmed training splits with a resulting mean token length of 142 tokens roughly matching the 140 token mean of the DAGPap22 dataset. From these passages 1,000 items were randomly sampled (with replacement) for each model found in Table \ref{tab:datasetstats}. Passages that were constructed using BLOOM and GPT-2 proceeded following the approach of \citep{liyanage_benchmark_2022} where the first sentence of the real passage was used as the prompt to construct the synthetic passage, subsequent generations were used to re-prompt the model to sample passages between 2 and 10 sentences. The first sentence from the real passage was then removed. For these models greedy decoding with a temperature of 1.0 was used. For SCIgen, 1,000 papers were generated and then a random passage of between 2 and 10 sentences was extracted from each one. For the paraphrase models, a randomly sampled passage from the human written passages were sent through a paraphrase tool. For the translation models, a human written passage was sent through the translation tool into a target language and then back translated into english. For all models generations, text similarity was measured between the original passage and the synthesized example, if the sample was more than 10\% similar it was not kept. This does simplify the problem and make the data less realistic as it removes synthetic passages that have a high lexical overlap with reference passages which might be common with inapporiate uses such as masking plagarism. The final dataset consisted of 110,474 passages of which 99,989 (90.5\%) were human written. This was done to try to match the extreme class imbalance that has been observed on synthetic scientific papers in the wild \citep{cabanac_prevalence_2021}. The final dataset was split by 80\%/10\%/10\% into train, validation, and test sets.

\section{Training details on models used}
\label{models}

For this work, all of our classification models were trained by finetuneing DeBERTa v3 large \citep{he_debertav3_2021} using the following hyperparameters: adamW optimizer, learning rate of 6e-6, batch size of 8, weight decay of 0.01 with warmup steps of 50. All classification models were trained for 3 epochs. For the domain adaptive pretraining (DAPT) model, we further pretrained using the parameters mentioned above with a masked language modeling objective on the Pubmed train split from the scientific papers dataset \citep{cohan_discourse-aware_2018} using 128 token chunks for 5 epochs. For the task adaptive pretraining (TAPT) model, we used the same approach with 5 epochs on the DAGPap22 dataset. Details of the SciBERT and logistic regression TF-IDF model baselines were not made available at the time of writing this paper. 

\end{document}